\documentclass[12pt]{article}
\usepackage{graphicx}
\usepackage{subfigure}
\usepackage{fullpage}
\usepackage{amsmath,amssymb}
\usepackage[ruled,lined]{algorithm2e}
\usepackage{color}
\usepackage{comment}
\usepackage{epstopdf}

\newcommand{\argmax}{\mathop{\mathrm{arg\,max}{}}}

\newtheorem{proposition}{Proposition}

\title{Thompson Sampling in Dynamic Systems for Contextual Bandit Problems}
\author{Tianbing Xu, Yaming Yu, John Turner, Amelia Regan \\
University of California, Irvine
}
\begin{document}
\maketitle

\abstract {
We consider the multi-arm bandit problems in the time-varying dynamic system for rich structural features. 
For the non-linear dynamic model, we propose the approximate inference for the posterior distributions 
based on Lapalace Approximation. For the context bandit problems, Thompson Sampling is adopted based on the 
underlying posterior distributions of the parameters. More specifically, we introduce the discount decays on 
the previous samples' impact and analyze the different decay rates with the underlying sample dynamics. 
Consequently, the exploration and exploitation is adaptively trade-off according to the dynamics in the system. 
}

\section{Introduction}
Contextual Bandit Problems have recently become popular because of their wide applicability in on-line advertising and information hosting. In these problems, at time $t$, a user represented as a feature vector (user) $x_t \in \mathbb{R}^d$, arrives and we need to decide which ad or information should be presented -- therefore which arm(e.g. ad)
from the pool $A$ should be selected for the user. After making this decision (e.g. picking arm $a$),
a response reward (e.g. a click or not signal $y_t$) is receive. 
This process is online for time $t = 1 \dots T$, then we collect samples $\{ z_t = (x_t, y_t)\}_{t=1}^T$.
The problem to find the best arm selection policy to produce the sequence decisions
which will achieve the best reward.  That is, close enough to the best oracle policy which has the benefit of hindsight.
This is a classic exploration/exploitation dilemma; in which we must determine the degree to which we should explore the unknown
arms to obtain more knowledge about the underlying system;
and when we should exploit the known arms to achieve the best expected reward.

For multi-arm bandit problems, there are many established policies, such as $\epsilon$-greedy,
Upper Confidence Bound(UCB)(\cite{auer:egreedy}),
the Gittins index method(\cite{john:gittins}) and
Thompson Sampling{\cite{thompson:ts,li:ts}}.
In $\epsilon$-greedy, with certain probability, we randomly pick an arm; while in the remaining probability, we 
pick the arm greedily with the largest expected reward known so far. 
When picking an arm, UCB not only consider the mean of the rewards, but also the uncertainties 
described by the confidence interval.
For Gittins index method, the reward is maximized by always continuing the bandit with the largest value of "dynamic allocation index".
Thompson Sampling randomly draws each arm according to its probability of being optimal.
The idea is heuristic; however, asymptotic convergence has been proven for contextual bandits (\cite{may:ts}).
There are many industrial applications, such as display ads(\cite{li:ts}), and news recommendation(\cite{li:news}). 
These learn the underlying parameters \textit{online}; however, 
there are no explicit assumptions or model for the parameters for the dynamic system.

Existing work mainly assumes that the online samples are from an i.i.d distribution or that the distribution does not vary over time. However, in many real industrial applications such as ad allocation and news recommendation,
the underlying distributions are changing over time.
For example, in the ads ranking system in Facebook,
we try to recommend the most suitable ads to any facebook user with respect to revenue optimization objectives
in term of his profile and all the related historical activities and other account information.
It is a dynamic ecosystem, as new users arrive we know little about them; long time users however, have rich structural features though all their activities and features are changing over time.
The underlying regression parameters are estimated by inference and learning; but as users arrive dynamically and unpredictably,
we can not assume that the underlying parameters are static, instead they are evolving with time.

There are many proposed methods for learning and inference in dynamical systems.
Kalman Filter(\cite{kalman:kf}) is an algorithm that works recursively on input data to estimate the underlying states (parameters) for a linear dynamic system.
West et. al. (\cite{west:dyn1},\cite{west:dyn2}) provide a systematic study of Bayesian inference for time series dynamic model.
An example for dynamic logistic regression can be referred to (\cite{penny:dlr}). Similar to West and Harrison (\cite{west:dyn1}),
McCormick et. al. (\cite{mccormick:dlr}) also introduce a "forgetting" factor for the inference in the online dynamic system.

Our work investigates the computational aspects of Thompson Sampling for bandit problems
in time-varying dynamical systems; we derive the approximate inference for the posterior distribution of regresssion parameters;
we introduce the discount decay on the likelihoods of previous samples;
we analyze the connections between the sample dynamics and discount decay rates;
and consequently we explain the adaptive trade-off of exploration and exploitation with the underlying dynamics.

\begin{figure} [t]
\centering
\includegraphics [width = 0.3 \linewidth]{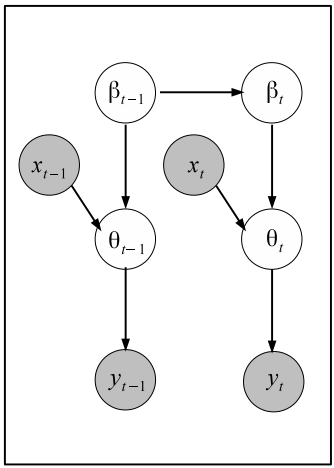}
\caption{Graphical Model for the Dynamic Contextual Model.}
\label{fig:dlr}
\end{figure}

\section{Dynamic Contexual Model}
\subsection{Model Description}
\label{sec:model}
At time $t$, $x_t \in \mathbb{R}^d$ arrives as the context feature,
with $y_t \in \{0,1\}$ the response reward such as click/conversion in online advertising.
Denote $D_t = D_{t-1} \cup z_t$  the samples untill time $t$ and $z_t = \{x_t,y_t\}$ is the observed sample at time $t$.

For each arm $a$, we denote $\theta_{t,a}$ the expected reward at time $t$;
$g_a$ the corresponding link function and $\beta{t,a}$ the regression parameter at time $t$. 
For notational simplicity, we just drop $a$ here for $\theta_{t,a}$ and $g_a$, $\beta_{t,a}$.
Suppose, at time $t$, each arm $a \in A$ has expected reward given by,
\begin{align*}
\theta_t=g(x_t, \beta_t) \;\;\;
\end{align*}
Here, each arm has a regression parameter $\beta_t \in \mathbb{R}^d$, and
if the parameter is a random variable, the reward is also random.

For the regression parameter, we introduce the dynamics from the random walk as,
\begin{align}
\label{eq:randwalk}
\beta_t = \beta_{t-1} + \epsilon_t, \qquad \epsilon_t \sim \mathcal{N} (0, Q_t)
\end{align}
here, $\epsilon_t$ is a white noise at time $t$ with covariance $Q_t$.
Different from the previous work on the regression parameters which are static and not varying with time;
here as time $t$ goes on, the parameters are evolving with time and our Brownian Motion is the simplest assumption.

For the link function $g$ for our generalized linear model,
in the context of online advertising, we usually have, logistic regression as,
\begin{align}
g=\frac{1}{1+exp(-x^T\beta)}
\end{align}
or probit regression,
\begin{align*}
g=\Phi(x^T\beta)
\end{align*}
where $\Phi$ is normal cdf, and $\theta_t = g(x_t, \beta_t) \in [0,1]$.
In this paper, we mainly focus on the nonlinear logistic function due to its popularity in industrial applications;
while the probit regression is a linear system and easier to handle.

Consequently, the reward is a sample from bernoulli trial,
$y_t \sim \mbox{Bernoulli}(\theta_t)$;  it is an observation that arises after an arm is chosen. 

More completely, in Figure(\ref{fig:dlr}), we formulate our dynamic contexual model for each arm as follows
for time $t = 1, 2, ... T$.
\begin{align*}
\beta_t = \beta_{t-1} + \epsilon_t, \qquad \epsilon_t \sim \mathcal{N} (0, Q_t) \\
\theta_{t} = g(x_t, \beta_t) \in [0,1], \qquad x_t \in \mathbb{R}^d \\
y_t \sim \mbox{Bernoulli} (\theta_t) \in \{0,1\}
\end{align*}
The prior distribution for the parameter is,
\begin{align*}
\beta_0 \sim \pi_0   , \qquad \beta_0 \in \mathbb{R}^{d} \\
\pi_0 = \mathcal{N}(u_0,\Sigma_0) \\
u_0 \in \mathbb{R}^d,\qquad \Sigma_0 \in \mathbb{R}^{d\times d}, \Sigma_0 \succ 0
\end{align*}

Once we decide which arm to pick at time $t$, the likelihood for sample  $z_t =\{x_t, y_t\}$ is,
\begin{align}
P(y_t|\beta_t, x_t)= \theta_t^{y_t} (1-\theta_t)^{1-y_t}, \qquad y_t \in \{0,1\}
\end{align}

Before time $t$, we have the historical samples $D_{t-1}$; at time $t$, we receive
features $x_t$, and would like to calculate the posterior distribution of regression 
parameter $\beta_t$.
Without knowing the sample $z_t = \{x_t, y_t \}$, we have the prior for $\beta_t$
as $P(\beta_t | D_{t-1})$. 
After the observation of $z_t$, we have information $D_t = D_{t-1} \cup z_t$,
the posterior $\pi_t(\beta_t|D_t)$ is updated online as,

\begin{equation*}
\pi_t(\beta_t|D_t) \propto P_t(\beta_t|D_{t-1}) P(y_t|\beta_t,x_t, D_{t-1})
\end{equation*}

As time progresses, additional samples arrive.
The updating process for the posterior distributions from time $1$ to $t$ is,
\begin{equation*}
\pi_0 \xrightarrow{z_1} \pi_1 \xrightarrow{z_2}  \pi_2 \dots \xrightarrow{z_t} \pi_t
\end{equation*}

\subsection{Thompson Sampling for the dynamic model}

Here we outline the $a$ in the related notation.
At time $t$, for each arm $a$, the parameter is $\beta_{t,a}$, the expected reward is $\theta_{t,a}$,
different arms have different likelihood functions.

At time t, Thompson Sampling picks an arm in proportion to the probability of being optimal.
Let $\omega_{t,a}$ denotes the probability of arm $a \in A$ having the highest expected reward.
\begin{align}
\label{eq:tsis}
\omega_{t,a}=P \{ {a}=\argmax_{a' \in A} \theta_{t,a'} | D_t\}
	= \mathbb{E} [1_{a}(\beta_{t,a})|D_t] \nonumber \\
\end{align}
here $1_{a}(\beta_{t,a})$ is 1 if arm $a$ has the highest expected reward; 
and the expectation is w.r.t $\pi_{t}(\beta_{t,a}|D_t)$, the posterior distribution of $\beta_{t,a} | D_t$ at time $t$.

To calculate the probability of being optimal, Thompson Sampling simply draws a sample for the parameter
of each arm from the posterior distribution and evaluates all the arms' corresponding expected reward.
Finally we pick the arm with the largest expected reward.

\section{Approximate Inference}
The dynamic context model is a non-linear system; for tractability, we estimate the posterior distribution using a Laplace approximation
in the closed form of the Gaussian distributions for the parameters at each time.
Here, we derive the recursive updating rules from time $t-1$ to $t$ for the regression parameter's posterior distribution.
At time $t-1$, the posterior distribution of $\beta_{t-1} | D_{t-1} \sim \mathcal{N}(u_{t-1}, \Sigma_{t-1})$,
that is,
\begin{align}
\label{eq:prevpost}
\pi_{t-1} = P(\beta_{t-1} | D_{t-1}) = \mathcal{N} (u_{t-1}, \Sigma_{t-1})
\end{align}
where, $u_{t-1}$ and $\Sigma_{t-1}$ are the posterior mean $E[\beta_{t-1} | D_{t-1}]$ and
posterior covariance $Cov[\beta_{t-1} | D_{t-1}]$ at time $t-1$.

From the random walk (Eq.\ref{eq:randwalk}), we have the prior at time $t$, $\beta_t | D_{t-1}$ as,
\begin{align}
\label{eq:prior}
P_t(\beta_t | D_{t-1}) = \mathcal{N} (u_{t|t-1}, \Sigma_{t|t-1})
\end{align}
here the prior mean $E[\beta_t | D_{t-1}] = u_{t|t-1} = u_{t-1}$,
and the prior covariance,
\begin{align}
\label{eq:priorVar}
Cov[\beta_t |  D_{t-1}] = \Sigma_{t|t-1} = \Sigma_{t-1} + Q_t
\end{align}

At time $t$, user $x_t$ arrives and a response reward $y_t$ observed after he makes decision. 
Now after observation of current training sample $z_t=\{x_t, y_t\}$, the posterior at time $t$, $\beta_t | D_t$ is,
\begin{equation}
\label{eq:recupdate}
\pi_t(\beta_t|D_t) \propto P_t(\beta_t|D_{t-1}) P(y_t|\beta_t,x_t, D_{t-1})
= P_t(\beta_t | D_{t-1}) P(y_t | x_t, \beta_t)
\end{equation}
here, $P_t(\beta_t | D_{t-1})$ is the parameter's prior distribution for time $t$, and $P(y_t | x_t, \beta_t)$ is the likelihood
of parameter given sample $z_t = \{x_t, y_t\}$.

The posterior $\pi_t$ is not in closed Gaussian form so here we adopt a Laplace Approximation.
By approximation, $\beta_t \sim \mathcal{N} (u_t, \Sigma_t)$; the posterior mean is
$E[\beta_t | D_t] = u_t$ and posterior covariance $Cov[\beta_t | D_t] = \Sigma_t$.

The log of the posterior is (drop the constant term),
\begin{align*}
L_t(\beta_t) = \log P(\beta_t | D_{t-1}) + \log P(y_t | x_t, \beta_t ) \nonumber \\
= - \frac{1}{2} (\beta_t - u_{t|t-1})^T (\Sigma_{t|t-1})^{-1} (\beta_t - u_{t|t-1})
  + y_t \log(\theta_t(x_t, \beta_t)) + (1 - y_t) \log(1 - \theta_t(x_t, \beta_t))
\end{align*}

The posterior mode $u_t$ is,
\begin{align*}
\nabla L_t(\beta_t) |_{u_t} =0 \Rightarrow \nonumber \\
\frac{\partial L_t(\beta_t)}{\partial \beta_t}
= H_{t|t-1}(\beta_t - u_{t|t-1}) + (y_t- \theta_t(x_t, \beta_t))x_t = 0 \nonumber \\
\end{align*}
Consequently, let $u_t = \beta_t$ in the above equation, the mean update is,
\begin{align}
\label{eq:meanupdate}
u_t = u_{t|t-1} + \Sigma_{t|t-1}(y_t- \hat{\theta}_t) x_t
\end{align}
where $\hat{\theta}_t$ is the approximated expected reward.
In order to have a clean closed form, we have the approximation for
$\theta_t(x_t, u_t)$ as,
\begin{align*}
\hat{\theta}_t \approx \theta_t(x_t, u_{t|t-1}) 
\end{align*}
and the Hessian is the negative inverse of the corresponding covariance, e.g.
$H_{t|t-1} = - (\Sigma_{t|t-1})^{-1}$.

Now, the Hessian matrix is the second derivative at the posterior mode,
\begin{align*}
H_t= \frac{\partial ^2 L(\beta_t)}{\partial \beta_t \partial \beta_t^T}|_{\beta_t= u_t}
\end{align*}

Consequently, the Hessian update is,
\begin{align}
\label{eq:hessian}
H_t = H_{t|t-1}-\hat{\theta}_t(1-\hat{\theta}_t) x_t \cdot x_t^T
\end{align}

By the Sherman Morrison formula(\cite{sherman:inv}) for the matrix inverse, the corresponding covariance is,
\begin{align}
\label{eq:covariance}
\Sigma_t = \Sigma_{t|t-1} - \frac{\hat{\theta}_t (1 - \hat{\theta}_t)} {1 + \hat{\theta}_t(1 - \hat{\theta}_t) s_t^2 }
(\Sigma_{t|t-1} x_t)(\Sigma_{t|t-1} x_t)^T
\end{align}
where, $s_t^2 = x_t \Sigma_{t|t-1} x_t^T$ is the variance of the activation $x_t^T \beta_t | D_{t-1}$.

Here, $\hat{\theta}_t (1 - \hat{\theta}_t)$ is approximated as the variance of Bernoulli prediction, and if we choose this
variance of $y_t$ as the observation noise in the Kalman Filter(\cite{kalman:kf}), the updates for the mean and
covariance matrix(Eqs.\ref{eq:meanupdate},\ref{eq:covariance}) is similar to Extended Kalman Filter.
In fact, these recursive updates are the same as the Iterative Extended Kalman Filter(\textbf{IEKF}).

\section{Explanations for Discount Decay}

In a dynamic system, samples from earlier periods have diminishing impacts on the current prediction. For example,
people may adopt time windows to train samples over a certain range of time. But how long should the time window be?
In this section, under a specific scenario, we introduce the explanations for the discount decay for the
previous sample's impacts on the current prediction. Furthermore, an analysis of decay rates and bounds are provided.

\subsection{The Discount Decay}
In many realistic scenarios, the impact of the previous samples has a decaying discount on the current posterior;
such as, for the sample $z_t$ at time $t$, the likelihood would have a decaying impact for the posterior $\pi_T$ 
at current time $T$ as a function decreasing as $T$ increases.
For example, the far away the current time $T$ to the sample $z_t$, the smaller the impact is.

Here, we introduce a special scenario to explain the decay functions.
For the approximate inference, the covariance $\Sigma_{t | t-1}$ is updated as in Eq.(\ref{eq:priorVar}),
\textit{specifically}, if we have $Q_t = q_t \Sigma_{t-1}$ (\cite{mccormick:dlr},\cite{west:dyn1}),
\begin{align}
\Sigma_{t | t -1} = (1 + q_t) \Sigma_{t-1} = \frac{1} {\lambda_t} \Sigma_{t-1}
\end{align}
here, $\lambda_t = \frac{1} { 1 + q_t}$. Assume $q_t > 0$, $0 < \lambda_t \leq 1$.
In the trivial case, when $\lambda_t = 1$, we have $Q_t = 0$, then there is no dynamics;
it is the conventional logistic regression. 

For $\lambda_t < 1$, the corresponding Hessian (observed information) is reduced as $H_{t | t-1} = \lambda_t H_{t-1}$;
and the increases with larger covariance,
there is discount on the information obtained from inference based on the previous observations.
More formally, from Eqs.(\ref{eq:prior},\ref{eq:prevpost}), for parameter $\beta$, as $u_{t|t-1} = u_{t-1}$,
we have the connection between the prior distribution at time $t$ and the posterior distribution at $t-1$ as,
\begin{align*}
P_t(\beta | D_{t-1}) = \mathcal{N} (u_{t-1}, \frac{1} {\lambda_t} \Sigma_{t-1})
= {\pi_{t-1} (\beta | D_{t-1})}^{\lambda_t}
\end{align*}

According to the recursive update for the posterior at time $t$ in Eq.(\ref{eq:recupdate}), we have,
\begin{align*}
\pi_t(\beta | D_t) \propto {\pi_{t-1} (\beta | D_{t-1})}^{\lambda_t} \ell_t(z_t | \beta)
\end{align*}
where, $\ell_t(z_t | \beta) = P (y_t | x_t, \beta)$ the likelihood on the sample $z_t$.
By induction based on the above recursive update, dropping the constant terms,
for samples from $t = 1$ to $T$, we get,
\begin{align}
\label{eq:decayupdate}
\pi_T(\beta | D_T) \propto \pi_0^{\lambda_{0:T}} \prod_{t=1}^{T} {\ell_t(z_t)^{\lambda_{t:T}}}
\end{align}
here $\lambda_{t:T}$ is the discount factor for sample $z_t$'s likelihood $\ell_t(z_t)$ untill time $T$, and it is,
\begin{align}
\lambda_{t:T} = \prod_{\tau = t+1}^T \lambda_{\tau}
\end{align}
and $\lambda_{T:T} = 1$.
The discount factor $\lambda_{t:T}$ is a cumulative factor depends from time $t$ 
where the sample is, to current time $T$ where the posterior is updated.  

\subsection{Different Decay Rates}
Without loss of generality, we consider $\lambda_{1:T}$, the discount factor for the first sample's impact untill time $T$.  Then we perform an analysis of the discount decay rate and lower bounds.
Let us assume certain functional forms to characterize $q_t$, the covariance increase rate of consecutive distributions,
as power function and discount function families.

\textbf{Power Function Family}

For the power function
\begin{align*}
q_t = \eta t^{-p}, \eta > 0
\end{align*}
the discount factor is,
\begin{align}
\label{eq:powerdiscount}
\lambda_{1:T} = \prod_{t=2}^T \frac{1} {1 + \eta t^{-p}} = \exp\{- \sum_{t=2}^T {\log(1 + \eta t^{-p})} \}
\end{align}

It is not difficult to show, for $p \leq 0 $ or $ 0 < p \leq 1 $, $\sum_t {\log(1 + \eta t^{-p})}$ is unbounded,
thus $\lambda_{1:T} \to 0$ as $T \to \infty$; for $p > 1$,
$\sum_t {\log(1 + \eta t^{-p})}$ is upper bounded, and thus, there exists a lower bound.
Furthermore, we analyze the different decay rates under different conditions for $p$ in Appendix(\ref{sec:ap1}).

\textbf{Exponential Function Family}

For the exponential function
\begin{align*}
q_t = \eta \gamma^t, \eta > 0, \gamma > 0
\end{align*}
the discount factor is,
\begin{align}
\label{eq:expdiscount}
\lambda_{1:T} = \prod_{t=2}^T \frac{1} {1 + \eta \gamma^t } = \exp\{- \sum_{t=2}^T {\log(1 + \eta \gamma^t)} \}
\end{align}
If $\gamma \geq 1$, $\lambda_{1:T} \to 0$ as $T \to \infty$; if $0 < \gamma < 1$,
$\sum_t {\log(1 + \eta \gamma^t)}$ is upper bounded; and consequently, the discount factor is lower bounded. 
Furthermore, we analyze the different decay rates under different conditions for $\gamma$ in Appendix(\ref{sec:ap2}).

\begin{figure} [t]
\centering
\includegraphics [width = 0.8 \linewidth]
{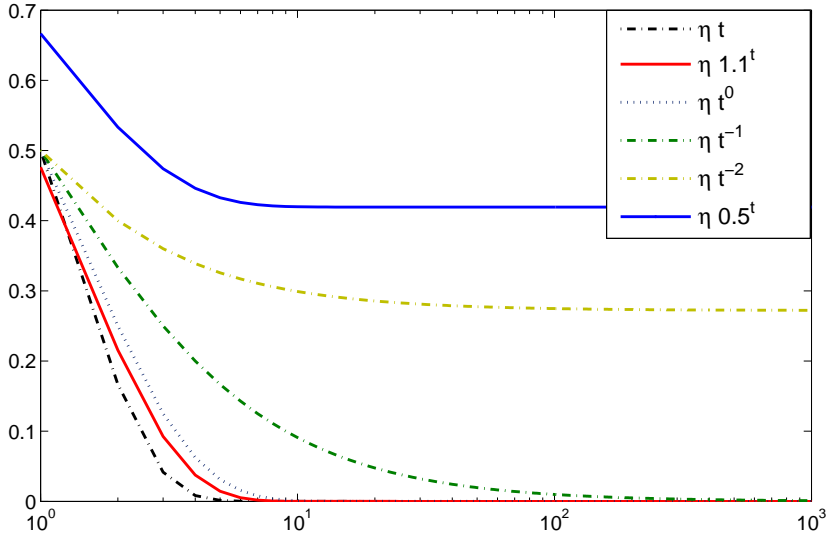}
\caption{Decay Rates for different power (dash curves) and exponential (solid curves) functions with $\eta=1$.
The decay rates in the figure are consist with the rates analysis.
x-axis is $T$, y-axis is discount factor $\lambda_{1:T}$.}
\label{fig:rates}
\end{figure}

Figure(\ref{fig:rates}) plots the different decay rates for the power function and exponential functions.
Finally, we summarize the behavior of the decay rate in the following Proposition.
\begin{proposition}
For the covariance update rate defined as power function and exponential function as above, with the infinite samples,
for the power function with $p \leq 1$ and exponential function with $\gamma \geq 1$,
the discount factor Eqs.(\ref{eq:powerdiscount},\ref{eq:expdiscount}) would be diminished;
for the $p > 1$ of power function and $0 < \gamma < 1$ of exponential function,
there exist corresponding lower bounds for the decay rates. 
With different parameter settings, the \textit{asymptotical} discount decay rates are on the order
from super-exponential, exponential and sub-exponential to power law rates.
\end{proposition}
Here, we list the detailed rates derived in Appendix(\ref{sec:ap})
\\
For power function, $\eta > 0$,
\begin{align*}
\lambda_{1 : T} \sim \left \{
\begin{array}{ll}
\exp\{pT \log T \},& \qquad p < 0 \\
\exp\{-(\log(1+\eta)) T \}, & \qquad p = 0 \\
\exp\{- \frac{\eta}{1-p} T^{1-p} \},& \qquad 0 < p < 1\\
(T + \eta)^{-\eta}, & \qquad p = 1 \\
\frac{\eta}{p-1} T^{1-p},& \qquad p > 1\\
\end{array}
\right.
\end{align*}
For exponential function, $\eta > 0$ and $\gamma > 0$,
\begin{align*}
\lambda_{1 : T} \sim \left \{
\begin{array}{ll}
\frac{\eta \gamma}{ 1 - \gamma} \gamma^T,& \qquad 0 < \gamma < 1 \\
\exp\{-(\log(1+\eta)) T \}, & \qquad \gamma = 1 \\
\exp \{-(\log \gamma) T^2 \},& \qquad \gamma > 1 \\
\end{array}
\right.
\end{align*}
For the rapid change of sample distributions in the dynamic system, it is straight forward to see,
the discount factor would quickly decay to zero, and there would be no impacts on the further prediction;
then the previous samples could be safely dropped.
Otherwise, if we have a slower and more smooth evolution of sample distribution,
there is still a lower bound on the discount decay, all the previous samples
still have effects on the further prediction no matter how far away the future time.

Considering the lower bounds of the decay rates, as $T \to \infty$, 
for power function family, when $p > 1$,
\begin{align*}
\lambda_{1:T} > \exp \{ -\frac{\eta}{p-1}\}, \qquad p > 1, \eta > 0
\end{align*}
for exponential function family, when $0 < \gamma < 1$,
\begin{align*}
\lambda_{1:T} > \exp\{-\frac{\eta}{1 - \gamma} \},
\qquad 0 < \gamma < 1, \eta > 0
\end{align*}

\textbf{Example 1}
A simple example is the constant increase at each time as $q_t = \eta > 0$.
Then we have the exponential decay rate $\lambda_{1:T} = (1 + \eta)^{1-T}$.
As $T \to \infty$, the rate is convergent to $0$.

\textbf{Example 2}
The second example is $q_t = \frac{\eta} {t^2}$, the increase rate is decreasing with time and
would approach to $0$. Then the covariance is bounded and
we have the power decay rate $\lambda_{1:T} \propto \exp \{-\eta (1 - \frac{1} {T}) \}$.
As $T \to \infty$, the rate is convergent to $\exp\{-\eta\}$.

\section{Experiments}

Here we use simulations to investigate how different exploration and exploitation policies work
in the dynamic system. First, Thompson Sampling is compared with $\epsilon$ greedy and UCB1 (\cite{auer:egreedy}).
based on the same approximate inference for the dynamic system.
In addition, for our dynamic simulation data, to investigate the performance gains,
we use Thompson Sampling to do the comparison between the logistic regression and our dynamic logistic regression.

\subsection{Simulation Setup}
The simulation here is similar to the process described in the simulation study of (\cite{may:ts}).
The number of actions $|A|$ is set at 10.
We randomly generated data set with time window $T = 5000$ from the model definition of the subsection(\ref{sec:model}).

More specifically, for each arm $a$, at time $t$, the parameter is,
\begin{align}
\label{eq:beta}
\beta_{t, a} \sim \mathcal{N} \{\beta_{t-1,a}, Q_t\}
\end{align}
The covariance of the regressors are either stationary or non-stationary according to our definition of $Q_t$.

The samples $\{x_t, y_t\}$ for each arm are generated sequentially as follows.
At time $t = 1 \dots T$, with feature dimensions $d = 10$,
\begin{itemize}
\item{sample feature vector $x_t \sim U(-1,0.5)^d$}
\item{sample true regressor $\beta_{t,a}$ according to Eq.(\ref{eq:beta})}
\item{calculate the true generalized linear regression expected reward(e.g.CTR) $\theta_{t,a} = g_a(x_t^T \beta_{t,a})$ }
\item{sample the true click signal $y_{t,a} \sim \mbox{Bernoulli}(\theta_{t,a}) $ for each arm $a \in A$}
\end{itemize}

The exploration/explotation experiment is set up as the following for each arm $a \in A$.
\begin{itemize}
\item{Learn the mean of the regressor ${u}_{t,a}$ online at time $t$ according to Eq.(\ref{eq:meanupdate})}
\item{Calculate the estimated reward (CTR) ${\hat{\theta}}_{t,a} = g_a(x_t^T {u}_{t,a})$  }
\item{Sample the potential estimated reward $\hat{y}_{t,a}$ for each arm }
\item{Make the decision to pick arm $a_t$ based on the different action selection policies}
\item{Record the selected estimated reward $\hat{\theta}_{t,a_t}$ and $\hat{y}_{t,a_t}$ }.
\item{The best reward from Oracle is $\theta_t^* = \max_{a \in A}{\theta_{t,a}}$ based on true parameters from hindsight,
and the true reward is sampled as $y_t^* \sim \mbox{Bernoulli}(\theta_t^*)$ }
\end{itemize}

Furthermore, to simulate the special decay case, we set $Q_t = \delta_t \Sigma_{t-1}$,
and $\Sigma_0 = 0.001 I_d$, thus during the model generating process,
the covariance matrix is shrinked as $\Sigma_t = (1 + \delta_t)*\Sigma_{t-1}$.
For the approximate inference with $Q_t = q_t \Sigma_{t-1}$, we adopt model selection procedure as follows.
\begin{itemize}
\item {For the same modeling distribution, we generate 6 data sets, each with $T=5000$ samples}
\item {We randomly pick 5 data sets, run with different parameter settings for $q_t$, (e.g. different
$\eta$ and $p$ for power function $q_t = \eta t^{-p}$), and pick the paremters with the best average reward from
the 5 rewards (Eq.(\ref{eq:reward}))}
\item{Finally, we run different policies (e.g. Thompson Sampling) on the 6-th data set with the best parameter for $q_t$
and report the results.}
\end{itemize}

We mainly use reward ratio and average regret as the performance metrics for different policies.
The reward ratio is defined as,
\begin{align*}
Reward (T) = \frac{\sum_{t=1}^T {\hat{y}_{t,a_t }}} { \sum_{t=1}^T  y_t^* }
\end{align*}
To have better assessment of Reward Ratio, we need to reduce the variance for the sampling of both true and estimated click signal
by directly using the expected reward as (\cite{li:ts}),
\begin{align}
\label{eq:reward}
Reward (T) = \frac{\sum_{t=1}^T {\hat{\theta}_{t,a_t}}} {\sum_{t=1}^T {\theta}_t^* }
\end{align}
For the infinite sample size $T$, finally, the estimated reward would approach the best Oracle reward and thus,
$Reward(T) \to 1$ as $T \to \infty$.

Similarly, the average regret is computed as follows,
\begin{align}
\label{eq:regret}
Regret(T) = \frac{\sum_{t=1}^T ({\theta_{t}^* - \hat{\theta}_{t,a_t} })} {T}
\end{align}
As $T \to 0$, the average regret would approach $0$.
Actually, the average regret and reward ratio are the equivalent performance metrics.

\subsection{Experimental Results}

We generate two different data sets according to our different definitions of $\delta_t$.
One is stationary with $\delta_t = \frac{0.1} {t^2}$ as the covariance matrix is bounded from above;
the other is non-stationary with $\delta_t = \frac{0.1} {t}$ as the covariance matrix is unbounded.

During the inference for function $q_t$, by assumption, it is a power function.
For the general setting, we set $q_t = \frac{\eta}{t}$; as in the simulation, we have finite
samples and in the modeling, the power function is either $\frac{\eta} {t}$ or $\frac{\eta} {t^2}$;
which will be showed in experiments that there is not too much difference for finite samples.
From model selection, we pick the best parameters for $q_t$ based on 5 data sets, and it is used in the approximate inference.
Consequently, we report the comparison results in terms of reward ratio and average regret.
Additionally, for Thompson Sampling, we also do the comparison between logistic regression and the dynamic
logistic regression for our dynamic data sets. The logistic regression is the special case in
Eqs(\ref{eq:meanupdate},\ref{eq:covariance}) with $Q_t = 0$.

\begin{figure} [t]
\centering
\includegraphics [width = 0.8 \linewidth]
{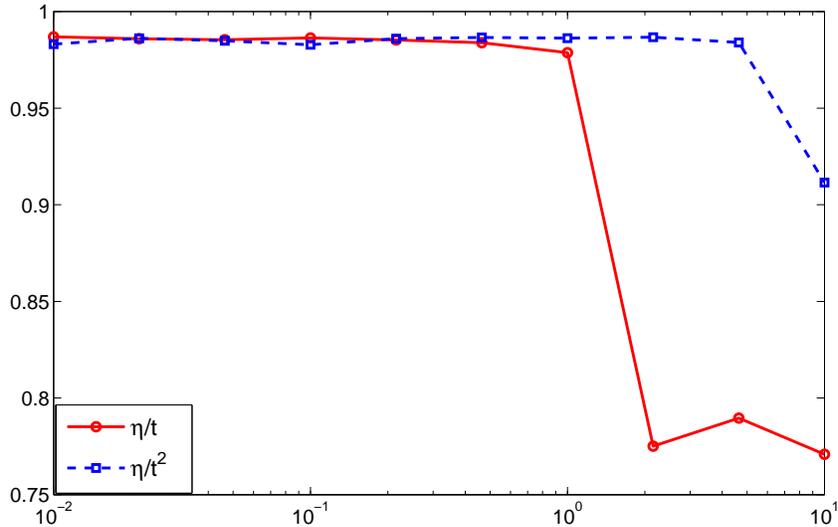}
\caption{Model selection Rewards for stationary data set.
Reward Ratio for different parameters $\eta$ of two power functions $\frac{\eta} {t}$
and $\frac{\eta}{t^2}$. x-axis is $\eta$, y-axis is reward ratio.}
\label{fig:seta}
\end{figure}

\begin{figure} [t]
\centering
\includegraphics [width = 0.8 \linewidth]
{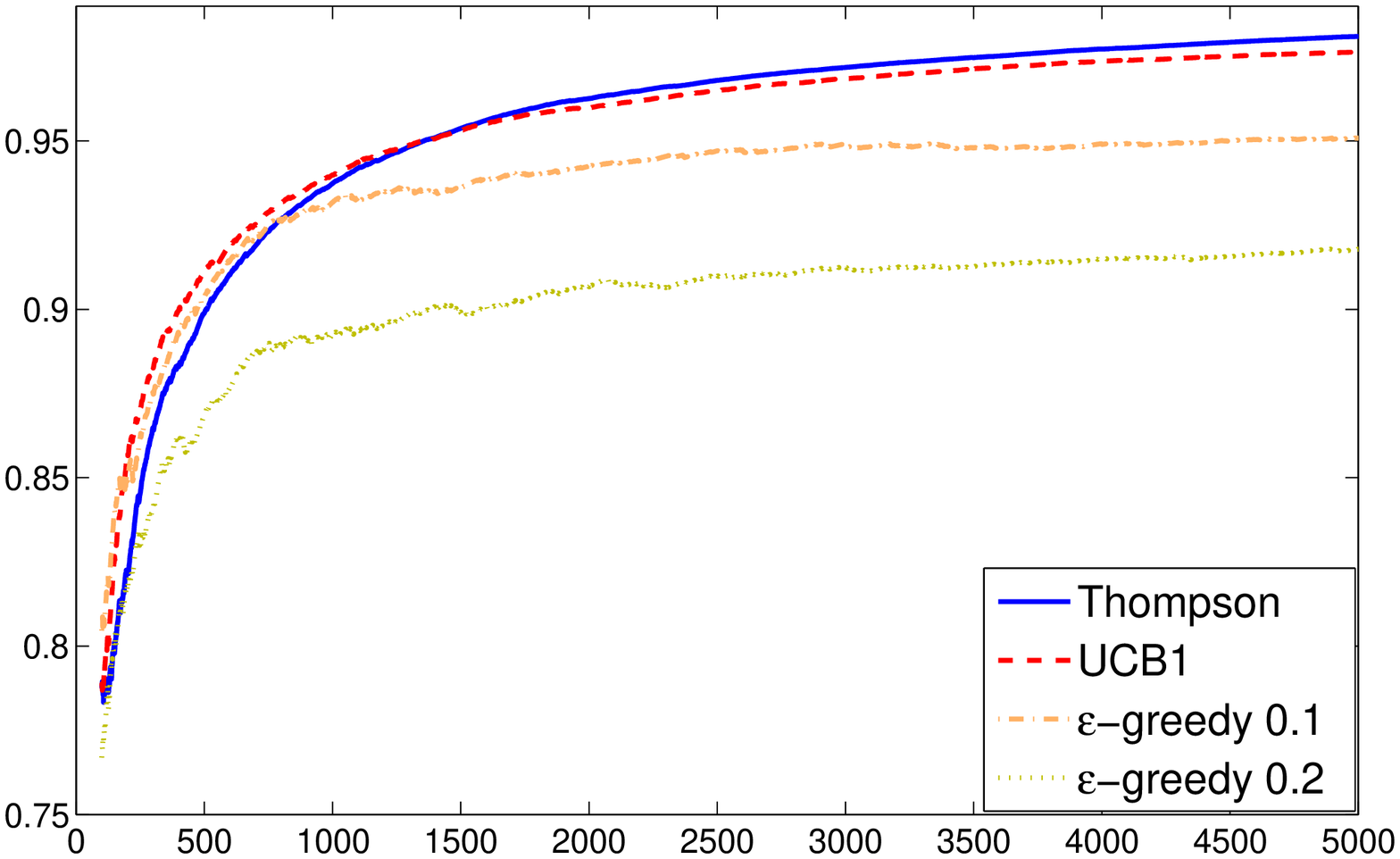}
\caption{Reward Ratio for stationary data set inferenced by $q_t = \frac{\eta} {t}$.
Thompson Sampling is compared with $\epsilon$-greedy and UCB1.
x-axis is $T$, y-axis is reward ratio.}
\label{fig:sreward}
\end{figure}

\begin{figure} [t]
\centering
\includegraphics [width = 0.8 \linewidth]
{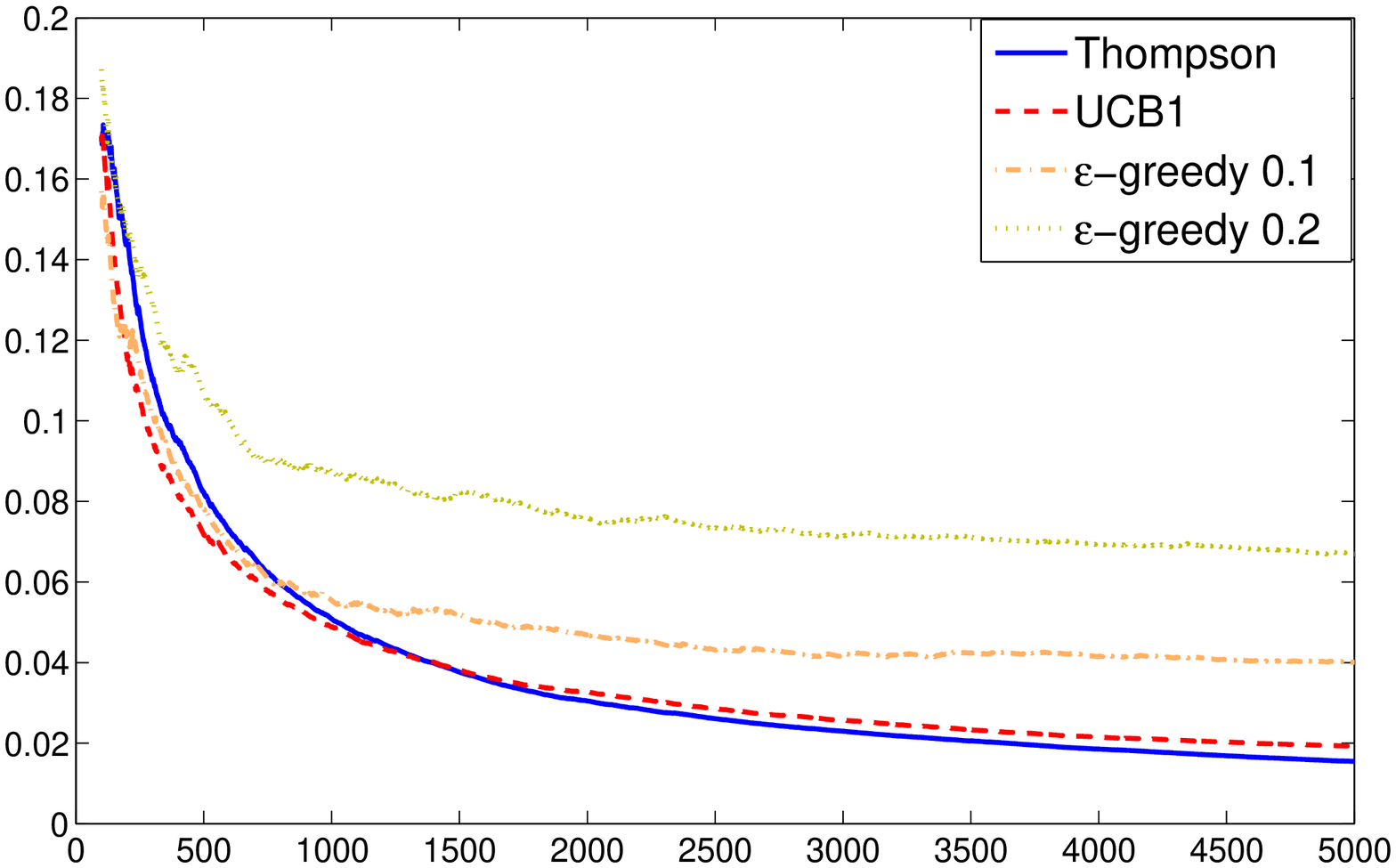}
\caption{Average Regret for stationary data set inferenced by $q_t = \frac{\eta} {t}$.
Thompson Sampling is compared with $\epsilon$-greedy and UCB1.
x-axis is $T$, y-axis is regret.}
\label{fig:sregret}
\end{figure}

\begin{figure} [t]
\centering
\includegraphics [width = 0.8 \linewidth]
{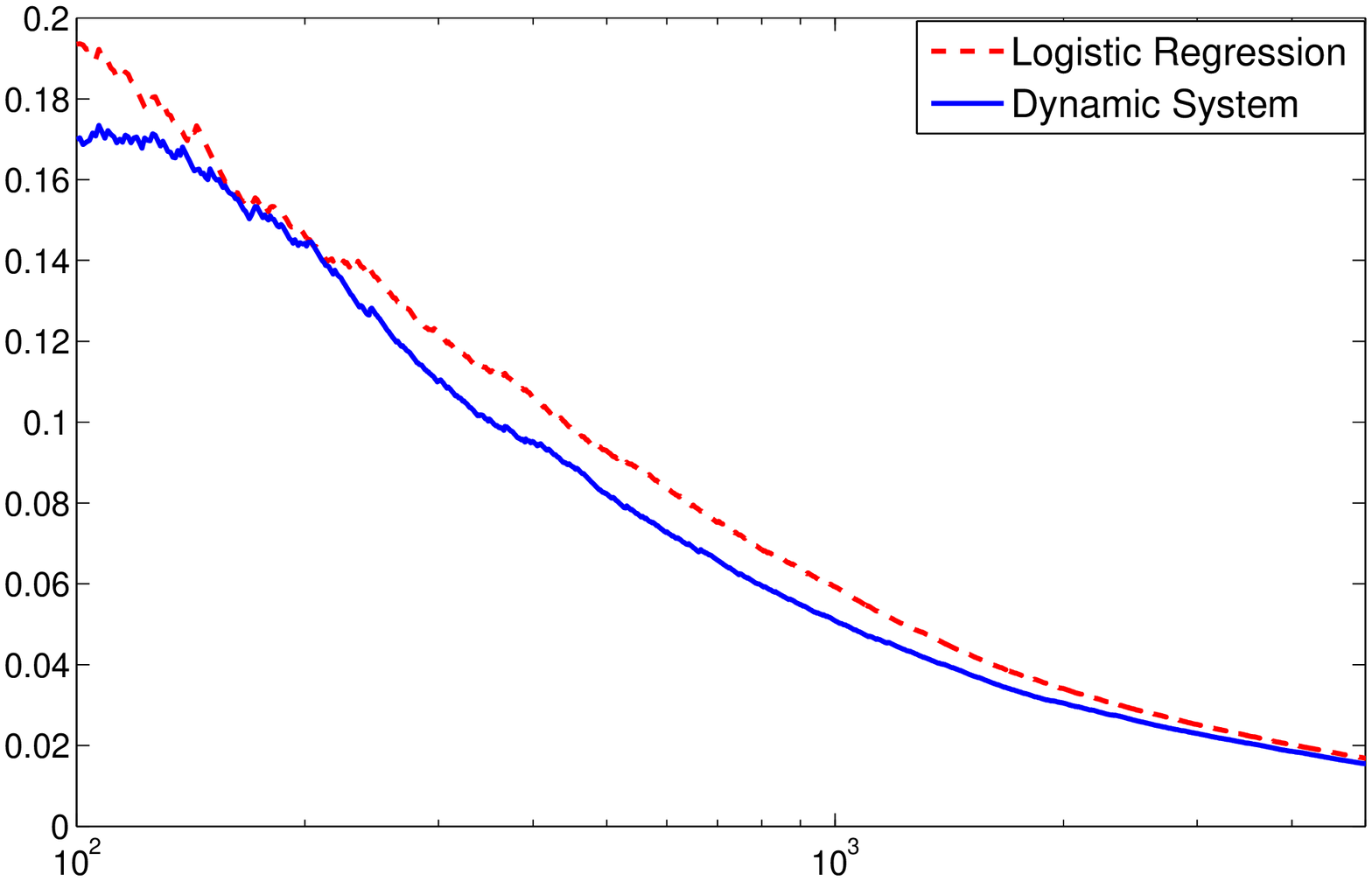}
\caption{Regrets for stationary data set by Thompson Sampling.
The dynamic logistic regression is compared with logistic regression for the same
stationary data.
x-axis is $T$, y-axis is regret.}
\label{fig:sdynamics}
\end{figure}

\textbf{Stationary Distribution}

Here we set $\delta_t = \frac{0.1}{t^2}$, it is not difficult to show that the covariance matrix
is bounded from above converges to a constant covariance matrix $\Sigma^*$. Then for each arm,
for sufficient large time $T$, the parameter is from the stationary distribution,
as $\beta_{T,a} \sim \mathcal{N} (u_0, \Sigma^*)$, with constant prior mean $u_0$.

Figure(\ref{fig:seta}) reports the model selection of different $\eta$ for two power functions. As it only has
finite samples and these two functions are not so different,
the correpsonding two best average rewards has little difference, but the function $\frac{\eta}{t^2}$
has a larger range of insensitive parameters choice as it decays slower than function $\frac{\eta}{t}$.

In Figures(\ref{fig:sreward},\ref{fig:sregret}), the comparison results for different exploration and exploitation policies
are illustrated. All these policies are based on the same approximate inference in dynamic system with $q_t = \frac{0.01}{t}$.
Thompson Sampling has the highest reward and lowest regret. The difference between Thompson Sampling and UCB1 is marginal;
this is a consist observation also in the simulation Study(\cite{may:ts}) as here the underlying parameters are
from \textit{asymptotical} stationary distribution.
And $\epsilon$-greedy is worse than Thompson Sampling and UCB1 as it does some portion of blind random search.

We also compare the Thompson Sampling in logistic regression to our dynamic logistic regression for stationary data. As expected,
in Figure(\ref{fig:sdynamics}), there is no really obvious difference after some point, since
\textit{asymptotically}, the parameters are from the same stationary distribution. Here, we do not report the comparison
result in term of reward as it is equivalent to the regret metric.

\begin{figure} [t]
\centering
\includegraphics [width = 0.8 \linewidth]
{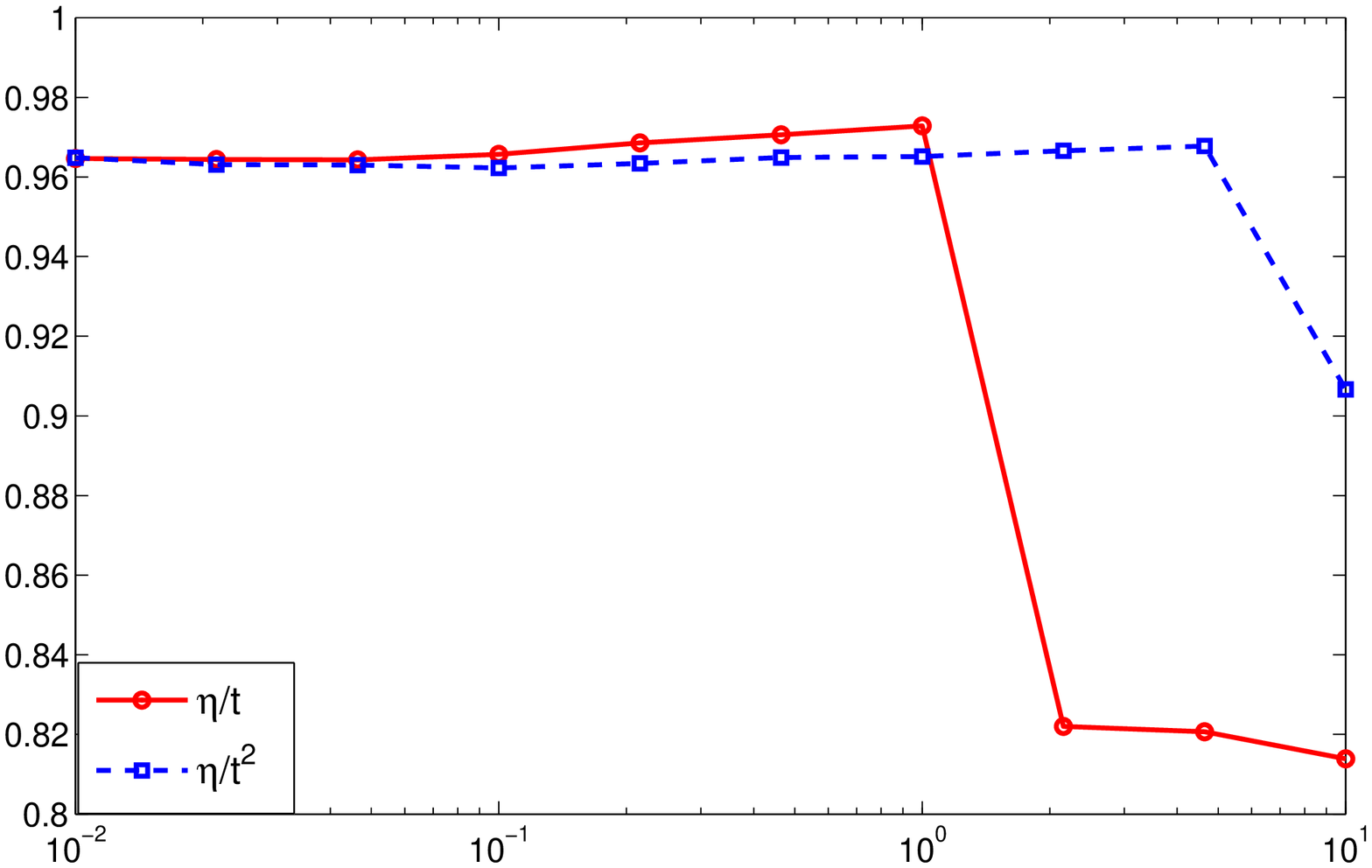}
\caption{Model selection Rewards for non-stationary data set.
Reward Ratio for different parameters $\eta$ of two power functions $\frac{\eta} {t}$
and $\frac{\eta}{t^2}$, used to do model selection. x-axis is $\eta$, y-axis is reward ratio.}
\label{fig:nseta}
\end{figure}

\begin{figure} [t]
\centering
\includegraphics [width = 0.8 \linewidth]
{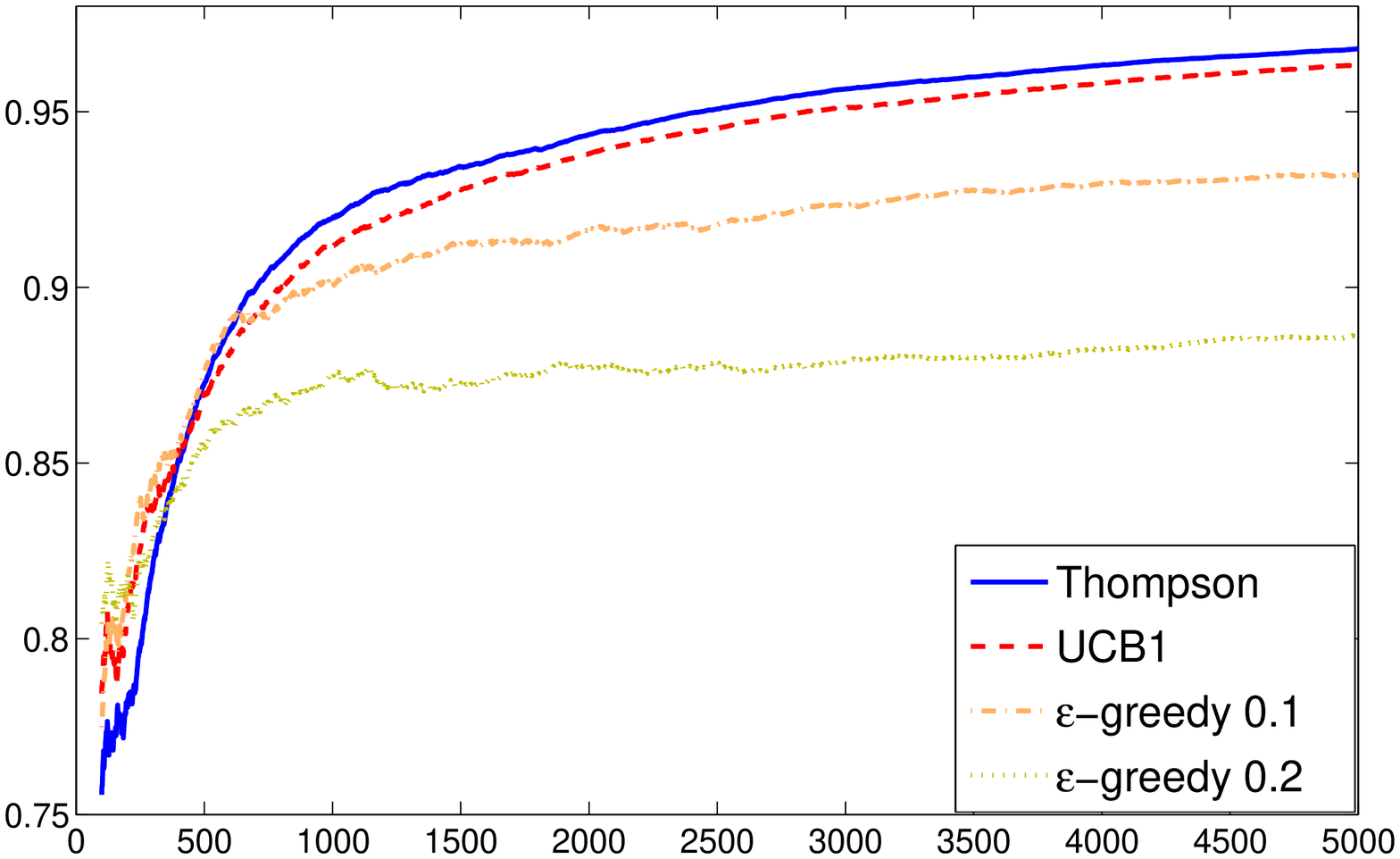}
\caption{Reward Ratio for non-stationary data set inferenced by $q_t = \frac{\eta} {t}$.
Thompson Sampling is compared with $\epsilon$-greedy and UCB1.
x-axis is $T$, y-axis is reward ratio.}
\label{fig:nsreward}
\end{figure}

\begin{figure} [t]
\centering
\includegraphics [width = 0.8 \linewidth]
{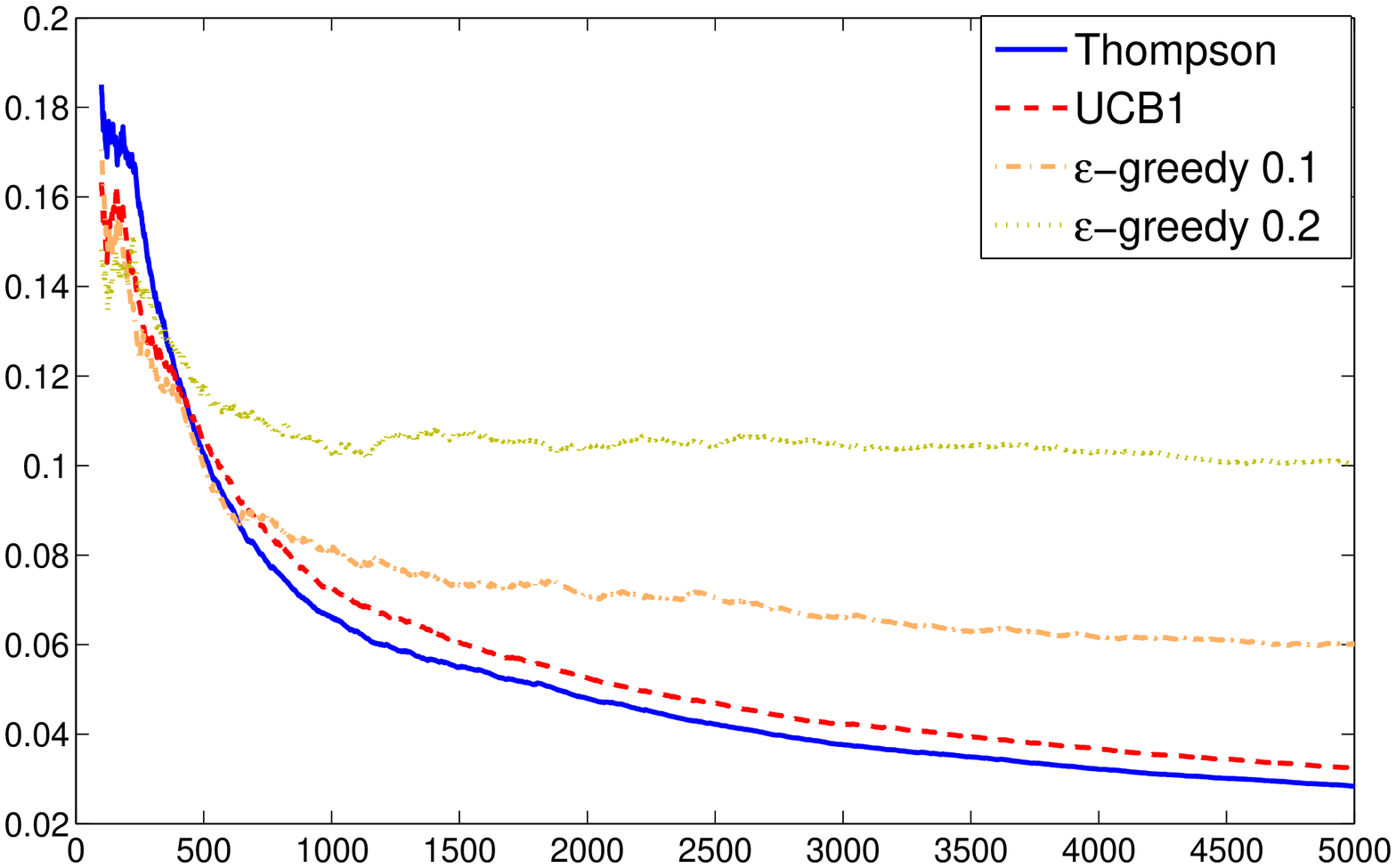}
\caption{Average Regret for non-stationary data set inferenced by $q_t = \frac{\eta} {t}$.
Thompson Sampling is compared with $\epsilon$-greedy and UCB1.
x-axis is $T$, y-axis is regret.}
\label{fig:nsregret}
\end{figure}

\begin{figure} [t]
\centering
\includegraphics [width = 0.8 \linewidth]
{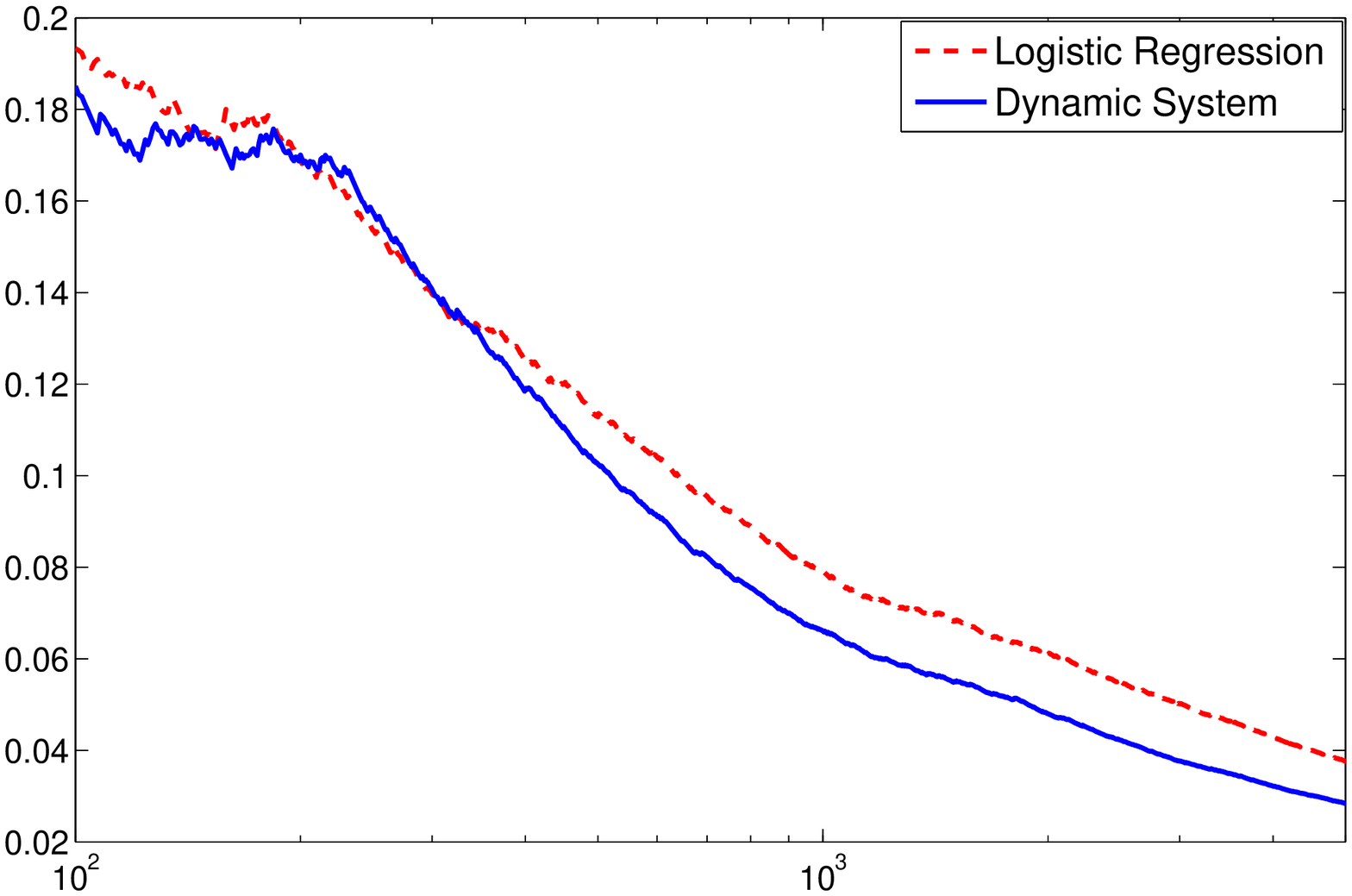}
\caption{Regrets for non-stationary data set by Thompson Sampling.
The dynamic logistic regression is compared with logistic regression for the same
non-stationary data.
x-axis is $T$, y-axis is regret.}
\label{fig:nsdynamics}
\end{figure}

\textbf{Non-Stationary Distribution}

Here, for the modeling, we set $\delta_t = \frac{0.1} {t}$. We may show that the covariance matrix for parameter
generated from this model will be unbounded; thus the samples are non-stationary.
Figure(\ref{fig:nseta}) illustrate the parameter choice for power function $\frac{\eta}{t}$ and $\frac{\eta}{t^2}$.
Obviously, $\frac{\eta}{t}$ has better average reward as the data is generated from this function form.
As same as for stationary distribution, $\frac{\eta}{t^2}$ has larger range for the parameter choice as it decays slower.
In figures(\ref{fig:nsreward},\ref{fig:nsregret}), we find that Thompson Sampling is better than UCB1 with larger margin.
It is mainly because that the approximate inference for the non-stationary samples are quite accurate; and the data are non-stationary.
As same as before, $\epsilon$-greedy is much more worse.
For the comparison between logistic regression and dynamic logistic regression in our non-linear and non-stationary
dynamic system, in Figure(\ref{fig:nsdynamics}) it ends up better result by dynamic logistic regression as
our approximate inference is able to adapt to the dynamics of the underlying distributions; while logistic regression assumes that the parameters are static and stationary.

\section{Discussion and Further work}

Let's take a close look at the posterior mean and covariance update again.

From Eq.(\ref{eq:meanupdate}), compared to logistic regression, we have,
\begin{align*}
u_t = u_{t-1} + \frac{1}{\lambda_t} \Sigma_{t-1} g_t, \,\,\
g_t \in \frac{\partial \log P(z_t|\beta_t))} {\partial \beta}
\end{align*}
here, $\frac{1} {\lambda_t} = 1 + q_t$ is the adaptive step size for the stochastic
gradient descent for mean $u_t$. This step size is adaptive to the underlying change of
parameter distribution characterized by $Q_t = q_t \Sigma_{t-1}$.

For the Hessian updates from Eq.(\ref{eq:covariance}),
\begin{align*}
H_t = \lambda_t H_{t-1}-\hat{\theta}_t(1-\hat{\theta}_t) x_t \cdot x_t^T
\end{align*}
Compared to logistic regression, the hessian(observed information) is reduced by $\lambda_t = \frac{1} {1 + q_t} < 1$
according to underlying distribution changes $Q_t$.

Thus, our approximate inference is able to adapt to exploration and exploitation trade-off.
When the sample distribution changes quickly($q_t$ in larger order), the discount factor decays quickly to 0,
and the observed information from the previous samples diminishes,
then it needs a larger step size to encourage \textit{exploration} in parameter space and consequently
search space for the expected posterior rewards with higher uncertainty.
If the sample distribution evolves smoothly ($q_t$ in smaller order),
the discount factor decays to the lower bound, information shrinkage is less heavy,
it ends up with smaller step size to encourage \textit{exploitation}.

For further work, we are interested in finding real applications with dynamic system for our Thompson Sampling.
In additional to learn $Q_t$, another possible extension is to study more complex dynamic system such as,
\begin{align*}
\beta_t = A_t \beta_{t-1} + \epsilon_t, \,\,\, \epsilon_t \sim \mathcal{N} (0,Q_t)
\end{align*}
and discuss the corresponding exploration and exploitation trade-off.

\appendix
\section{Decay Rates Analysis}
\label{sec:ap}
Here, we provide a detailed analysis of the discount decay rates under different power and exponential functions.
\subsection{Power Function Family}
\label{sec:ap1}
For the power function, the decay rate orders are discussed here for different $p$.
\\
\textbf{1. $p = 0$}.

Here, $\lambda_t = \frac{1} {1 + \eta} < 1$ is constant; the discount factor,
\begin{align}
\label{eq:expdecay1}
\lambda_{1 : T} =  (1 + \eta)^{-(T-1)} = (1 + \eta) \exp\{-(\log(1+\eta)) T \}, \qquad p = 0, \eta > 0
\end{align}
and $\log(1 +\eta) > 0$, thus, we have the \textit{exponential} decay rate.
\\
\textbf{2. $p = 1$}.

First, by the integration,
\begin{align*}
\int \log(1 + \frac{\eta}{t}) \, dt = \eta \log(t + \eta) + t \log(1 + \frac{\eta}{t})
\end{align*}
we have the upper bound,
\begin{align*}
\sum_{t=2}^T \log (1 + \frac{\eta}{t}) < \int_{1}^T {\log(1 + \frac{\eta}{t}) \, dt }
< \eta \log(T + \eta) + T \log(1 + \frac{\eta}{T})
\end{align*}
and the lower bound,
\begin{align*}
\sum_{t=2}^T \log (1 + \frac{\eta}{t}) > \int_{1}^T {\log(1 + \frac{\eta}{t}) \, dt } - \log (1 + \eta) \\
\end{align*}
Thus, the summation is in the order of,
\begin{align*}
\sum_{t=2}^T \log (1 + \frac{\eta}{t}) \sim
\eta \log(T + \eta) + T \log(1 + \frac{\eta}{T})
\end{align*}
Then, the decay rate is in the order,
\begin{align*}
\lambda_{1:T} \sim \exp\{-\eta \log(T + \eta) - T \log(1 + \frac{\eta}{T}) \}
, \qquad \eta > 0
\end{align*}
For sufficient large $T$, \textit{asymptotically}, we have,
\begin{align*}
T \log(1 + \frac{\eta}{T}) \to \eta
\end{align*}
Consequently, the \textit{asymptotical power law} decay rate order is,
\begin{align}
\label{eq:power1}
\lambda_{1:T} \sim \exp\{-\eta \log(T + \eta) \} = (T + \eta)^{-\eta} , \qquad p = 1, \eta > 0
\end{align}
\\
\textbf{3. $p > 0, p \neq 1$}.

First, we have the inequalities,
\begin{align*}
\eta t^{-p} - \frac{\eta^2}{2} t^{-2p} < \log (1 + \eta t^{-p}) < \eta t^{-p}
\end{align*}
Then, the summation is upper bounded by,
\begin{align*}
\sum_{t=2}^T \eta t^{-p} < \eta \int_{t=1}^T \, \frac{dt^{1-p}} {1 - p}
= \eta \frac{T^{1-p} - 1} {1-p}
\end{align*}
And lower bounded as,
\begin{align*}
\sum_{t=2}^T {\eta t^{-p} - \frac{\eta^2}{2} \sum_{t=2}^T t^{-2p}}
> \eta \int_{t=1}^T \frac{dt^{1-p}} {1 - p} - \eta - \frac{\eta^2}{2} \sum_{t=2}^T t^{-2p}
\end{align*}
Here, we have two cases on the second summation on the right hand side,
\\
$a)~ p = \frac{1}{2}$
\begin{align*}
\sum_{t = 2}^T t^{-2p} = \sum_{t = 2}^T \frac{1}{t} < \int_{t=1}^T \,d\log{t} < \log T
\end{align*}
\\
$b)~ p \neq \frac{1}{2}$
\begin{align*}
\sum_{t = 2}^T t^{-2p} < \int_{t=1}^T \, \frac{d{t}^{1-2p}} {1 - 2p} =  \frac{T^{1-2p} - 1}{1-2p}
\end{align*}
Then, the lower bound is,
\begin{align*}
\sum_{t=2}^T {\eta t^{-p} - \frac{\eta^2}{2} \sum_{t=2}^T t^{-2p}} >  \left \{
\begin{array}{lr}
\frac{\eta}{1-p} (T^{1-p} -1) - \eta - \frac{\eta^2}{2} \log T, & p = \frac{1}{2} \\
\frac{\eta}{1-p} (T^{1-p} -1) - \eta - \frac{\eta^2}{2(1-2p)} (T^{1-2p} - 1), & p \neq \frac{1}{2} \\
\end{array}
\right.
\end{align*}
Now, we have two different decay rates.
\\
\textbf{3.1 ~ $0 < p < 1$}
\\
For $p = \frac{1}{2}$, the discount rate order is,
\begin{align*}
\lambda_{1:T} \sim \exp\{-2\eta \sqrt{T} + \frac{\eta^2}{2} \log T \}
, \qquad \eta > 0
\end{align*}
as \textit{asymptotically}, $\log T = o(\sqrt{T})$, it is the \textit{sub-exponential} decay rate. \\
For $p \neq \frac{1}{2}$, we also has the \textit{sub-exponential} decay rate as,
\begin{align*}
\lambda_{1:T} \sim \exp\{- \frac{\eta}{1-p} T^{1-p} + \frac{\eta^2}{2(1-2p)} T^{1-2p} \}
, \qquad \frac{\eta}{1 - p} > 0
\end{align*}
Furthermore, the \textit{asymptotical sub-exponential} rate,
\begin{align}
\label{eq:subexp}
\lambda_{1:T} \sim \exp\{- \frac{\eta}{1-p} T^{1-p} \}, \qquad 0 < p < 1, \eta > 0
\end{align}
\\
\textbf{3.2 ~ $p > 1$}
\begin{align*}
\lambda_{1:T} \sim \exp\{ \frac{\eta}{p-1} T^{1-p} - \frac{\eta^2}{2(2p-1)} T^{1-2p} \} \\
\sim \exp\{ \frac{\eta}{p-1} T^{1-p} \}
, \qquad \frac{\eta}{p-1} > 0
\end{align*}
\textit{Asymptotically}, the \textit{power law} decay rate is in the order,
\begin{align}
\label{eq:power2}
\lambda_{1:T} \sim \frac{\eta}{p-1} T^{1-p}, \qquad p > 1, \eta > 0
\end{align}
Here, as $T \to \infty$, we have the lower bound for the discount decay as,
\begin{align}
\label{eq:powerbound}
\lambda_{1:T} > \exp \{ -\eta \frac{1 - T^{1-p}} {p - 1}\} \to \exp \{ -\frac{\eta}{p-1}\}, \qquad p > 1, \eta > 0
\end{align}
\\
\textbf{4. $p < 0$}.

First, we have,
\begin{align*}
\log (1 + \eta t^{-p}) = \log(\eta t^{-p}) + \log(1 + \frac{1}{\eta} t^p )
\end{align*}
The second term is equivelent to the case where $-p > 0$, \textit{asymptotically},
the summation is in the order of,
for $p \neq -1$,
\begin{align*}
\sum_{t=2}^T \log(1 + \frac{1}{\eta} t^p) \sim \frac{T^{1+p} - 1}{\eta(1+p)},\qquad \eta > 0, p < 0
\end{align*}
for $p = -1$,
\begin{align*}
\sum_{t=2}^T \log(1 + \frac{1}{\eta} t^{-1}) \sim \frac{1}{\eta} \log(T + \frac{1}{\eta}),
\qquad \eta > 0
\end{align*}
By the integration,
\begin{align*}
\sum_{t=2}^T \log t \sim \int_{t=1}^T \, d(t \log t -t) \sim T \log T - T
\end{align*}
then, for the first term, the summation,
\begin{align*}
\sum_{t=2}^T \log(\eta t^{-p}) = \sum_{t=2}^T \log \eta - p \sum_{t=2}^T \log t
\sim -p T \log T + pT +  T \log \eta
\end{align*}
Now, we show the \textit{asymptotically} decay rate is in the order of \textit{sup-exponential} as,
\begin{align}
\label{eq:supexp1}
\lambda_{1:T} \sim \exp\{pT \log T \}, \qquad p < 0
\end{align}
\textbf{4.1 ~ $ p = -1$}
\begin{align*}
\lambda_{1:T} \sim \exp\{ pT \log T -pT - T \log \eta - \frac{1}{\eta} \log (T + \frac{1}{\eta}) \}
\sim \exp\{ pT \log T \}
\end{align*}
\textbf{4.2 ~ $ p = -\frac{1}{2} $}
\begin{align*}
\lambda_{1:T} \sim \exp \{ pT \log T - pT - T \log \eta - \frac{T^{1+p}} {\eta(1+p)} \}
\sim \exp\{ pT \log T \}
\end{align*}
\textbf{4.3 ~ $0 < - p < 1, p \neq -\frac{1}{2}, -p > 1 $}
\begin{align*}
\lambda_{1:T} \sim \exp \{ pT \log T - pT - T \log \eta - \frac{T^{1+p} - 1} {\eta(1+p)} \}
\sim \exp\{ pT \log T \}
\end{align*}

\subsection{Exponential Function Family}
\label{sec:ap2}
For the exponential function we analyze the different decay rates under different conditions for $\gamma$.
\\
\textbf{1. $\gamma = 1$}

This case is the same as in the case 1 of power function, it has the same \textit{exponential} decay in Eq.(\ref{eq:expdecay1}).
\\
\textbf{2. $0 < \gamma < 1$}

First, we have the inequalities,
\begin{align*}
\eta \gamma^t - \frac{\eta^2}{2} < \log(1 + \eta \gamma^t ) < \eta \gamma^t
\end{align*}
and the upper bound,
\begin{align*}
\eta \sum_{t=2}^T \gamma^t < \eta \frac{1 - \gamma^{T+1}}{ 1 - \gamma}
\end{align*}
the lower bound,
\begin{align*}
\sum_{t=2}^T \log(1 + \eta \gamma^t) >  \eta (\frac{1 - \gamma^{T+1}}{ 1 - \gamma} - 1 - \gamma)
-\frac{\eta^2}{2} (\frac{1 - \gamma^{2(T+1)}}{1 - \gamma^2} - 1 - \gamma^2)
\end{align*}
as $\gamma^{2T} = o(\gamma^T)$, the discount decay is in the \textit{asymptotical} order,
\begin{align*}
\lambda_{1:T} \sim \exp \{ \frac{\eta \gamma}{1 - \gamma } \gamma^T  \}
\end{align*}
\textit{Asymptotoically}, we have the \textit{exponential} decay rate,
\begin{align}
\label{eq:expdecay2}
\lambda_{1:T} \sim \frac{\eta \gamma}{ 1 - \gamma} \gamma^T, \qquad 0 < \gamma < 1, \eta > 0
\end{align}
Here, as $T \to \infty$, we have a lower bound for the discount decay,
\begin{align}
\label{eq:expbound}
\lambda_{1:T} > \exp \{\eta \frac{\gamma^{T+1} - 1 }{1 - \gamma} \} \to \exp\{-\frac{\eta}{1 - \gamma} \},
\qquad 0 < \gamma < 1, \eta > 0
\end{align}
\\
\textbf{3. $ \gamma > 1$}

First, we have,
\begin{align*}
\log(1 + \eta \gamma^t) = \log (\eta \gamma^t) + \log(1 + \frac{1}{\eta} (\frac{1}{\gamma})^t ),
\qquad 0 < \frac{1}{\gamma} < 1
\end{align*}
\textit{asymptotically},  the summation of second term,
\begin{align*}
\sum_{t=2}^T \log(1 + \frac{1}{\eta} (\frac{1}{\gamma})^t ) \sim \frac{\gamma - \gamma^{-T}}{\eta(\gamma - 1) }
\end{align*}
the first term,
\begin{align*}
\sum_{t}^T \log (\eta \gamma^t) = \sum_{t=2}^T \log \eta + \log \gamma \sum_{t=2}^T t
= (T - 1) \log \eta + \frac{(T-1)(T+2)}{2} \log \gamma
\end{align*}
Then, the discount decay rate,
\begin{align*}
\lambda_{1:T} \sim \exp\{-(\log \gamma) T^2 - (\log \eta) T + \frac{\gamma^{-T}}{\eta(\gamma - 1)} \}
\end{align*}
Thus, \textit{asymptotically}, we have the \textit{super-exponential} order of decay rate,
\begin{align}
\label{eq:supexp2}
\lambda_{1:T} \sim \exp \{-(\log \gamma) T^2 \}, \qquad \gamma > 1
\end{align}

\newpage
\clearpage
\bibliography{dynts}
\bibliographystyle{alpha}
\end{document}